\documentclass[conference]{IEEEtran}
\IEEEoverridecommandlockouts

\usepackage{graphicx}
\usepackage{amsmath}
\usepackage{amssymb}
\usepackage{booktabs}
\usepackage{url}

\graphicspath{{figures/}}

\newcommand{\Xt}{\widehat{\mathbf{X}}}
\newcommand{\X}{\mathbf{X}}

\begin{document}

\title{Compressing AI Traffic: Standardized Neural Network Coding of
Visual-Token Representations in Split Vision--Language Inference}

\author{\IEEEauthorblockN{Reza Heidari}
\IEEEauthorblockA{
\textit{Aalto University, Nokia Technologies}\\
Espoo, Finland \\
reza.heidari@aalto.fi}
\and
\IEEEauthorblockN{Hamed R. Tavakoli}
\IEEEauthorblockA{
\textit{Nokia Technologies}\\
Espoo, Finland\\
hamed.rezazadegan\_tavakoli@nokia.com}
\and
\IEEEauthorblockN{Juho Kannala}
\IEEEauthorblockA{
\textit{Aalto University}\\
Espoo, Finland \\
juho.kannala@aalto.fi}
}

\maketitle

\begin{abstract}
When the visual encoder and the language decoder of a vision--language model
(VLM) run on different compute nodes, the intermediate visual-token embeddings
become a communicated payload rather than an internal activation. We call such
machine-consumed intermediate tensors \emph{AI traffic} and ask how far they
can be compressed with a \emph{standardized}, training-free codec. We insert
ISO/IEC 15938-17 Neural Network Coding (NNC) round trips on the complete visual
interface of a Qwen3-VL-8B-Instruct video question answering pipeline,
comprising the main visual-token representation and the DeepStack feature
streams, while leaving weights, prompts, and generation untouched, and sweep
the quantization parameter (QP) over a wide rate range. Closed-ended Video-MME
accuracy remains close to the uncompressed reference up to a
$98\%$ reduction of the transmitted BF16 tensor and only then collapses;
open-ended MLVU generation shows the same plateau-and-collapse profile under an
LLM judge. This robustness is not due to near-lossless reconstruction: the
decoded tensor is heavily discretized, carries substantial row-wise relative
$L_2$ error, and has a visibly steeper singular-value decay than its source.
Downstream reasoning therefore depends on coarse structure and relative geometry
rather than exact floating-point values, which argues for rate--task rather than
rate--distortion optimization of AI traffic codecs.
\end{abstract}


\section{Introduction}

Large vision--language models (VLMs) pair a visual encoder with a language
model: the encoder maps the visual input to a sequence of visual-token
embeddings, and the language model consumes those tokens with a text prompt to
produce an answer. Video inputs make this interface expensive, since a long
token sequence must carry spatial content, temporal change, event ordering,
and long-range context at once.

In edge and device--cloud deployments the encoder and decoder need not be
co-located: a device may run the visual encoder near the sensor while a much
larger decoder runs on a capable node~\cite{kang2017neurosurgeon}. Such a split
pipeline need not transmit the raw video at all---it can transmit the
intermediate visual tokens instead. The transmitted object is then not a video
bitstream but an intermediate AI representation, which we call \emph{AI
traffic}: visual tokens, feature maps, latent embeddings, attention states, or
any other tensor exchanged between AI modules. As models grow and spread over
heterogeneous hardware, reducing AI traffic becomes relevant for bandwidth,
latency, and system efficiency.

We ask how much visual-feature traffic in a VLM video question answering
pipeline can be compressed before downstream task performance degrades. To
answer this, we apply Neural Network Coding (NNC), standardized as
ISO/IEC 15938-17~\cite{kirchhoffer2022nnr}, to the complete visual interface
of Qwen3-VL~\cite{qwenvl}. This interface comprises the primary visual-token
representation together with the DeepStack feature streams injected into the
language model. Each stream is independently encoded to a bitstream, decoded,
and replaced by its reconstruction before language-model inference continues.

Two findings organize the paper. Visual-token embeddings are highly
compressible---very aggressive rate reduction is tolerated before performance
collapses. Yet the reconstructions are strongly discretized and substantially
distorted, and the decoder still operates well, so exact floating-point
reconstruction is unnecessary for downstream VLM reasoning provided enough
task-relevant structure survives.

Our contributions are: (i) we formulate visual-token transport in split VLM
inference as an AI traffic coding problem and instantiate it with a
standardized codec rather than a bespoke or learned quantizer; (ii) we report a
full rate--task sweep on closed-ended (Video-MME) and open-ended (MLVU) video
understanding with identity and zero-token controls that bound the useful
operating range from both sides; and (iii) we give a representation-level
analysis explaining \emph{why} strong compression is tolerated, showing that
tensor-domain distortion is a poor proxy for task utility in this regime.

\section{Related Work}

Contemporary video question answering is built on large VLMs, in which the
visual tokens form the interface between perception and generation. We use
Qwen3-VL-8B-Instruct~\cite{qwenvl}, a multimodal family supporting long
interleaved contexts, with video-specific mechanisms for spatio-temporal
modeling and temporal grounding.

Neural network compression---quantization, pruning, low-rank approximation,
entropy coding, distillation, weight sharing---is most often applied to model
parameters, but the same tools apply to intermediate tensors. NNC specifies a
compressed representation and decoding process for neural-network
data~\cite{kirchhoffer2022nnr}, with entropy coding based on
DeepCABAC~\cite{wiedemann2020deepcabac}. We use the NNCodec reference
software~\cite{becking2023nncodec}.

Compressing signals for machine rather than human consumption has been studied
as video coding for machines~\cite{duan2020vcm} and collaborative
intelligence, where deep features are compressed between device and
server~\cite{choi2018deepfeature,shao2020bottleneck}. That work generally
trains the compression stage for a specific downstream network. Our setting
differs: the payload is the visual-token interface of a general-purpose
generative VLM rather than a convolutional feature map; the consumer is an
autoregressive language decoder rather than a classifier; and the codec is a
fixed, standardized, training-free tensor coder applied at inference time.

\section{Compressing AI Traffic}

\begin{figure}[!t]
\centering
\includegraphics[width=\columnwidth]{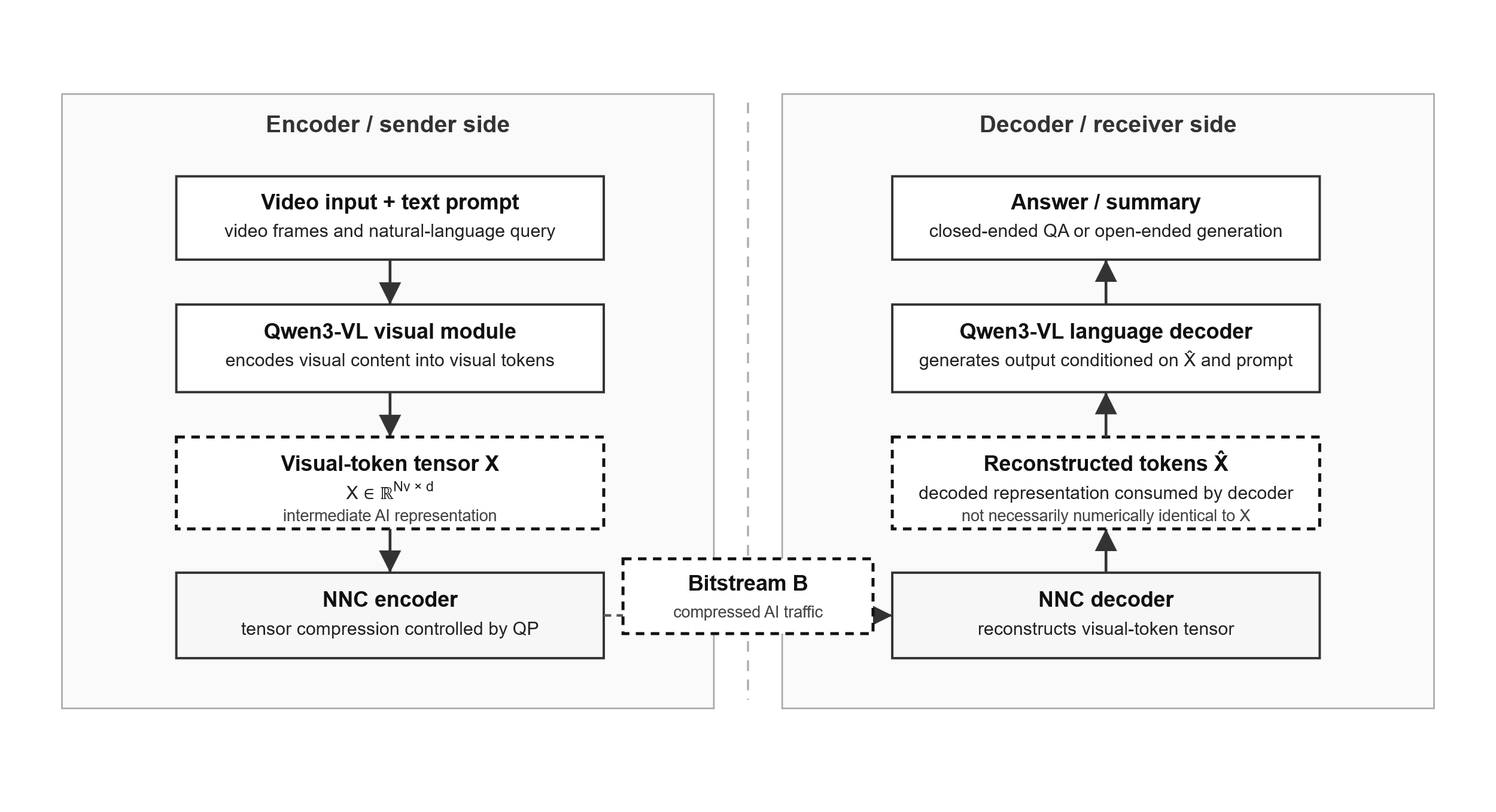}
\caption{NNC-based compression of visual-feature AI traffic. The vision module
emits a main visual-token representation together with DeepStack feature
streams. Each stream is independently NNC encoded, transmitted, decoded, and
substituted by its reconstruction before language-model inference.}
\label{fig:pipeline}
\end{figure}

AI traffic compression differs from conventional media coding in its success
criterion. Video compression targets a signal intended for human viewing, and
the decoded output is judged by perceptual or signal-level resemblance to the
original. AI traffic compression targets representations intended for machine
reasoning; the decoded output need not be visually meaningful and only has to
preserve what the receiving model requires.

Let the visual interface produced by the vision module be
\[
\mathcal{X} =
\left\{
\mathbf{X}_{\mathrm{main}},
\mathbf{X}_{\mathrm{DS},1},
\ldots,
\mathbf{X}_{\mathrm{DS},K}
\right\},
\]
where $\mathbf{X}_{\mathrm{main}}$ is the primary visual-token representation
and $\mathbf{X}_{\mathrm{DS},k}$ denotes the $k$-th DeepStack feature stream.
In Qwen3-VL-8B-Instruct, $K=3$. Each tensor is independently coded using the
same NNC operating point,
\[
B_i =
\mathrm{NNC}_{\mathrm{enc}}
\left(\mathbf{X}_i;\,\mathrm{QP}\right),
\qquad
\widehat{\mathbf{X}}_i =
\mathrm{NNC}_{\mathrm{dec}}(B_i),
\]
for $\mathbf{X}_i \in \mathcal{X}$. The reconstructed interface
\[
\widehat{\mathcal{X}} =
\left\{
\widehat{\mathbf{X}}_{\mathrm{main}},
\widehat{\mathbf{X}}_{\mathrm{DS},1},
\ldots,
\widehat{\mathbf{X}}_{\mathrm{DS},K}
\right\}
\]
replaces the original visual interface before language-model inference
continues. QP governs compression strength and is shared across all streams.

The evaluation question is whether $\widehat{\mathcal{X}}$ retains enough
task-relevant information for downstream question answering. Because the
reconstruction is consumed by a model rather than viewed by a human, downstream
task performance is the primary measure of useful reconstruction quality---a
reconstruction may have large mean squared error yet preserve the structure the
decoder needs, and conversely numerical accuracy does not guarantee that
task-relevant relations survive. We therefore report both task performance and
representation-level properties.

\section{Experimental Setup}

\textbf{Model and intervention.} We use Qwen3-VL-8B-Instruct in BF16 with
batch size $1$. The visual forward pass produces a primary visual-token
representation together with three DeepStack feature streams. We intercept
this complete visual interface and independently apply an NNC encode--decode
round trip to each of the four tensors. The reconstructed tensors are then
returned in place of their corresponding originals before language-model
inference continues. Weights, tokenizer, prompt format, language decoder, and
generation function are otherwise unmodified, so any change in output is
attributable to compression of the visual interface.

\textbf{Codec.} Each source tensor is detached, cast to FP32, and made
contiguous before encoding; each decoded tensor is cast back to the device and
dtype expected by the pipeline. NNC is configured with uniform approximation,
dependent quantization and row skipping enabled, quantize-only mode and TCA
disabled, and sparsity $0.0$; only QP varies across the sweep.

\textbf{Conditions.} \emph{Identity} passes the visual interface through
unchanged and gives the uncompressed reference. \emph{Zero-token} replaces it
with zeros and acts as a sanity check on how much the model depends on the
visual representation. \emph{NNC} substitutes the reconstruction.

\textbf{Data and metrics.} Closed-ended evaluation uses
Video-MME~\cite{fu2025videomme}: the model is prompted with the question and
options, the output is parsed into an option letter, and accuracy is computed
by strict matching. We also report the \emph{valid rate}, the fraction of
generations parseable into a valid option, which separates task errors from
format failures. Open-ended evaluation uses the summary subset of
MLVU~\cite{zhou2025mlvu}, scored by a GPT-4 judge that assigns completeness
and reliability out of $5$ each, summed to $10$. For both benchmarks, $100$
test samples are drawn after deterministic sorting, skipping samples whose
video files cannot be resolved. Compression is reported relative to the
aggregate BF16 size of the complete visual interface, i.e., the sum of the
primary visual-token and all DeepStack feature tensors, which is the
communication-relevant quantity in the split pipeline.

\section{Results}

\begin{figure}[!t]
\centering
\includegraphics[width=\columnwidth]{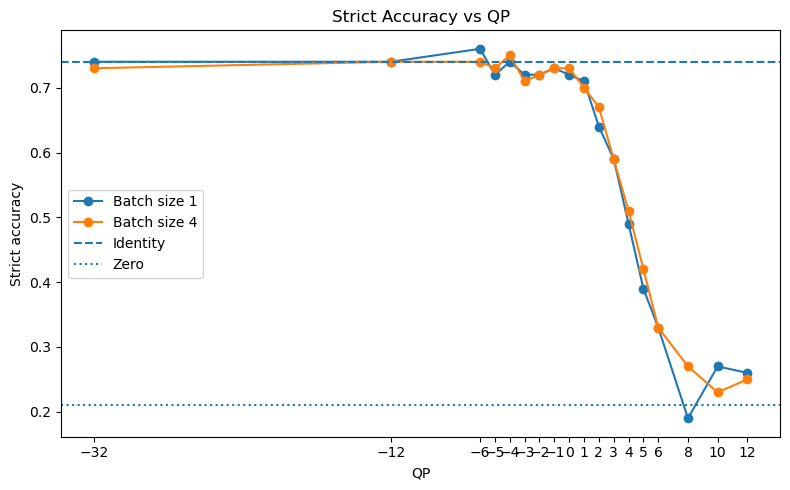}
\caption{Video-MME accuracy across QP. Accuracy tracks the identity baseline
over a wide QP range and drops only at aggressive compression. The zero-token
baseline is a visual-information sanity check.}
\label{fig:acc}
\end{figure}

\begin{table}[!t]
\renewcommand{\arraystretch}{1.05}
\caption{Video-MME results. Compression is relative to the BF16 source tensor.}
\label{tab:videomme}
\centering
\footnotesize
\setlength{\tabcolsep}{4pt}
\begin{tabular}{@{}llccc@{}}
\toprule
\textbf{Codec} & \textbf{QP} & \textbf{Valid rate} &
\textbf{Accuracy} & \textbf{Comp.\ (\%)} \\
\midrule
Identity & --  & 1.00 & 0.74 & --    \\
Zero     & --  & 0.55 & 0.21 & --    \\
\midrule
NNC & $-32$ & 1.00 & 0.74 & 58.41 \\
NNC & $-12$ & 1.00 & 0.74 & 87.41 \\
NNC & $-6$  & 1.00 & 0.76 & 94.09 \\
NNC & $-3$  & 1.00 & 0.72 & 96.45 \\
NNC & $0$   & 1.00 & 0.72 & 98.09 \\
NNC & $1$   & 1.00 & 0.71 & 98.68 \\
NNC & $2$   & 1.00 & 0.64 & 99.08 \\
NNC & $3$   & 1.00 & 0.59 & 99.35 \\
NNC & $4$   & 1.00 & 0.49 & 99.54 \\
NNC & $6$   & 0.86 & 0.33 & 99.87 \\
NNC & $12$  & 0.70 & 0.26 & 99.99 \\
\bottomrule
\end{tabular}
\end{table}

\textbf{Rate behavior.} Increasing QP monotonically shrinks the encoded
bitstream, confirming that QP is a usable rate control for visual-token
tensors. Compression relative to the BF16 source spans $58.4\%$ at
$\mathrm{QP}=-32$ to $99.99\%$ at $\mathrm{QP}=12$
(Table~\ref{tab:videomme}).

\textbf{Closed-ended accuracy.} Table~\ref{tab:videomme} and
Fig.~\ref{fig:acc} show a clear rate--task trade-off. Across low, moderate, and
even fairly high QP the NNC condition tracks the identity baseline, so $\Xt$
retains enough information for the decoder to answer many closed-ended
questions correctly; at higher QP accuracy falls, so aggressive compression
does eventually remove task-relevant information.

This rules out two trivial explanations. The model is not insensitive to the
visual representation: zeroing the tokens drops accuracy from $0.74$ to $0.21$.
And the useful operating region is not confined to near-lossless coding:
accuracy stays within a few points of the reference up to
$\mathrm{QP}=0$, where the bitstream is around $99\%$ smaller than the BF16
tensor. The valid rate confirms that the high-QP drop is not a formatting
artifact---parsing is perfect throughout the plateau and degrades only in the
collapse region, where the model also begins emitting unparseable output.

\begin{figure}[!t]
\centering
\includegraphics[width=\columnwidth]{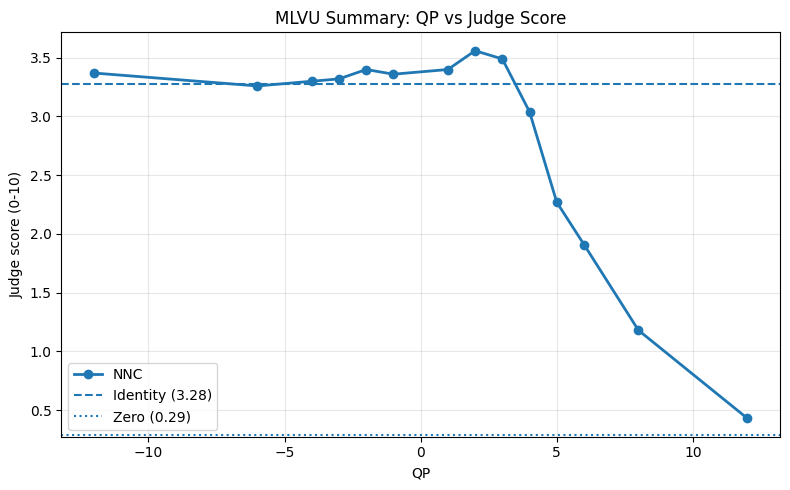}
\caption{MLVU judge score across QP, against identity and zero-token
references.}
\label{fig:mlvu}
\end{figure}

\textbf{Open-ended generation.} MLVU requires free-form output and therefore
tests whether compressed tokens support longer semantic generation.
Fig.~\ref{fig:mlvu} shows the same pattern: moderate compression preserves most
of the generation quality and only aggressive compression degrades it.
Agreement between a strict discrete metric and a free-form judged metric
indicates the plateau is not an artifact of multiple-choice answering, where a
model might guess the right option from weak visual evidence.

\textbf{Plateau and collapse.} Both benchmarks exhibit a
plateau-and-collapse profile: compression increases while performance holds
near the identity baseline, then a small additional rate reduction causes a
sharp drop. This is operationally useful, identifying a QP range in which
communication cost falls by one to two orders of magnitude at essentially no
task cost, and it implies a deployed system should be configured
conservatively relative to the knee rather than at it.

\section{Representation-Level Analysis}

For the representation-level analysis, we focus on the primary visual-token
stream and denote it by $\X=\mathbf{X}_{\mathrm{main}}$ and its reconstruction
by $\Xt=\widehat{\mathbf{X}}_{\mathrm{main}}$.

\begin{figure*}[!t]
\centering
\begin{minipage}[t]{0.47\textwidth}
    \centering
    \includegraphics[height=0.18\textheight,keepaspectratio]{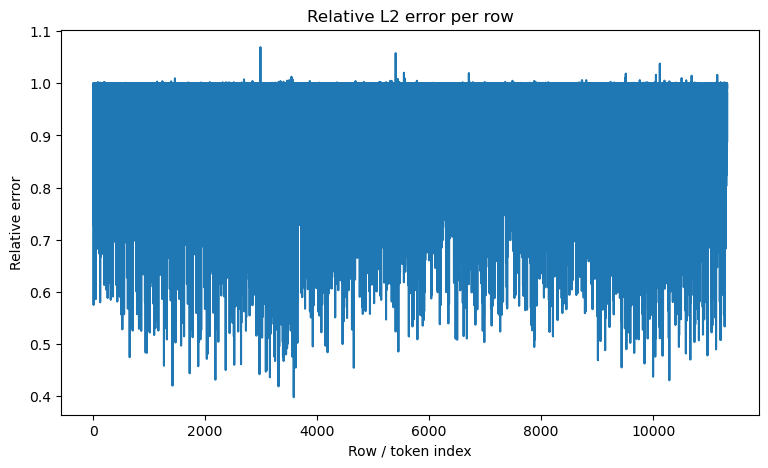}\\[-1mm]
    {\footnotesize (a) Row-wise relative $L_2$ error}
\end{minipage}
\hfill
\begin{minipage}[t]{0.52\textwidth}
    \centering
    \includegraphics[height=0.18\textheight,keepaspectratio]{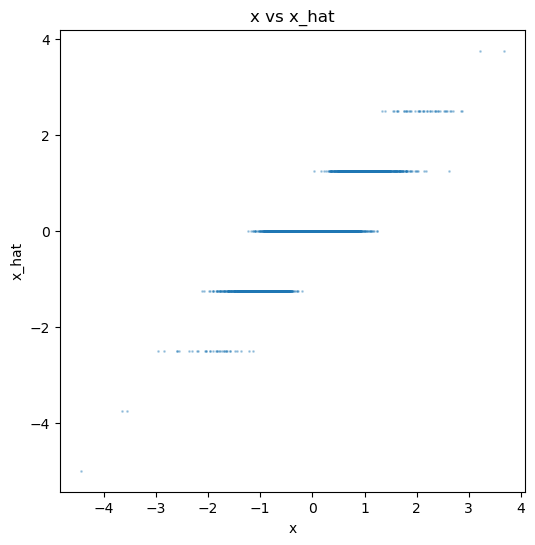}\\[-1mm]
    {\footnotesize (b) $\X$ versus $\Xt$}
\end{minipage}
\caption{Distortion of the primary visual-token representation after NNC
compression. (a) Row-wise relative $L_2$ error shows substantial numerical
distortion. (b) Continuous source values collapse onto discrete reconstruction
levels, revealing strong quantization.}
\label{fig:distortion}
\end{figure*}

Comparing $\X$ with $\Xt$ shows that the reconstruction is not near-identical,
particularly at stronger settings: row-wise relative $L_2$ error is substantial
(Fig.~\ref{fig:distortion}(a)). Downstream robustness is therefore \emph{not}
explained by near-lossless reconstruction.

\begin{figure}[!t]
\centering
\begin{minipage}[t]{0.49\columnwidth}
    \centering
    \includegraphics[width=\linewidth]{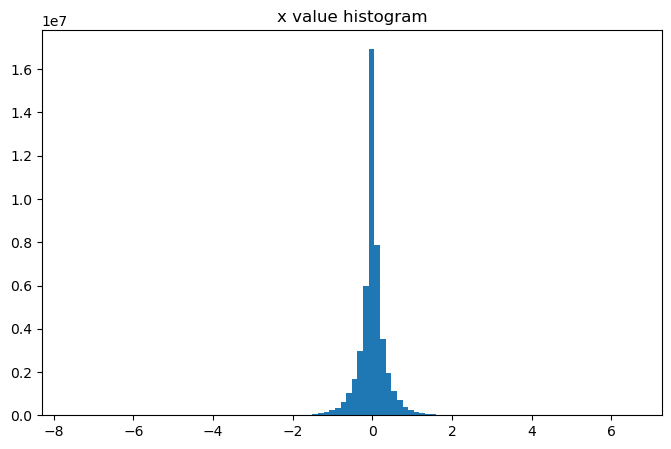}\\[-1mm]
    {\footnotesize (a) Source $\X$}
\end{minipage}
\hfill
\begin{minipage}[t]{0.49\columnwidth}
    \centering
    \includegraphics[width=\linewidth]{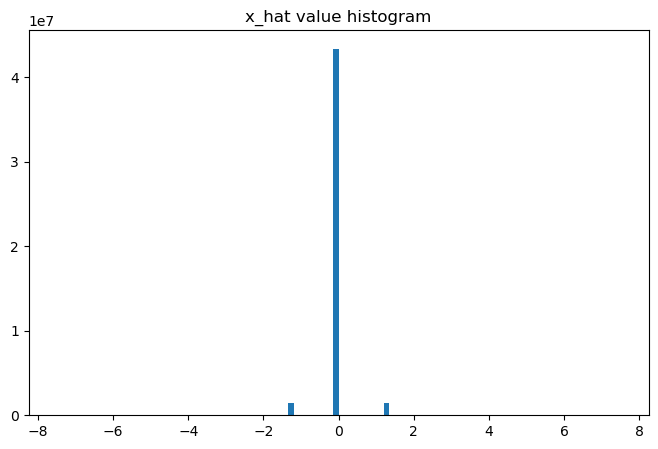}\\[-1mm]
    {\footnotesize (b) Reconstruction $\Xt$}
\end{minipage}
\caption{Value distributions of the primary visual-token representation before
and after NNC compression. The reconstruction concentrates on a smaller set of
discrete values, consistent with strong quantization.}
\label{fig:hist}
\end{figure}

Histogram analysis (Fig.~\ref{fig:hist}) separates the two tensors clearly:
the source has a comparatively smooth value distribution, whereas the
reconstruction concentrates on discrete levels. Scatter analysis
(Fig.~\ref{fig:distortion}(b)) reinforces this, showing continuous source
ranges collapsing onto quantized reconstruction levels. Despite this
substantial value-level distortion, downstream performance remains stable over
a wide QP range.

\begin{figure}[!t]
\centering
\includegraphics[width=\columnwidth]{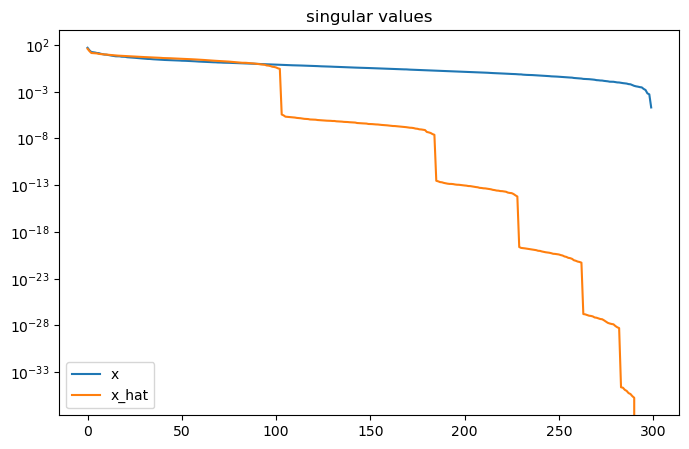}
\caption{Singular-value spectra of $\X$ versus $\Xt$. The steeper decay is
consistent with a reduction in effective rank.}
\label{fig:singular}
\end{figure}

Spectral analysis gives a third view. The reconstruction's singular values
decay more steeply than the source's (Fig.~\ref{fig:singular}), so compression
suppresses or removes lower-energy directions and reduces the effective
dimensionality of the representation. Since performance is stable in the
plateau region, many of those suppressed components appear non-essential for
the evaluated tasks.

\section{Discussion and Conclusion}

The visual interface of the tested pipeline contains a large amount of
compressible redundancy, but also task-relevant information that cannot be
discarded indefinitely, as the high-QP collapse shows. Because a heavily
discretized and spectrally simplified reconstruction still supports
high-quality output, tensor-domain distortion is a weak predictor of task
utility here; AI traffic codecs should be evaluated on rate--task rather than
rate--distortion curves. From a systems standpoint, compressing the
visual-token interface reduces communication cost without architectural
changes or retraining, and using NNC supplies a standardized tensor-coding
pipeline rather than an ad hoc quantizer---which matters for interoperability
between components from different vendors.

The study is limited to one backbone and subsets of two benchmarks, so the
observed behavior may depend on architecture, tokenization, task, and
interception point. Open-ended evaluation also relies on an LLM judge.
Future work should broaden models and tasks, measure encoding/decoding and
end-to-end latency, and explore adaptive rather than uniform rate allocation.

In summary, visual-token AI traffic in a split vision--language pipeline can be
reduced by well over an order of magnitude---up to $98\%$ relative to the
BF16 source tensor---with essentially no loss in downstream task performance,
after which performance collapses sharply. The reconstructed tokens are not
exact numerical copies, yet they retain enough task-relevant structure for the
language decoder to operate effectively.


\end{document}